\pdfoutput=1
\documentclass[11pt]{article}
\usepackage{graphicx}
\usepackage{amsmath}
\usepackage{booktabs}
\usepackage{multirow}
\usepackage{enumitem}
\usepackage[table]{xcolor}
\usepackage{pifont}
\usepackage{booktabs, colortbl, xcolor}
\usepackage{nicematrix}
\usepackage{tikz} 
\usepackage[preprint]{acl}

\usepackage{times}
\usepackage{latexsym}

\usepackage[T1]{fontenc}
\usepackage[utf8]{inputenc}

\usepackage{microtype}

\usepackage{inconsolata}

\usepackage{subcaption}
\title{PATS: Process-Level Adaptive Thinking Mode Switching}

\author{
\textbf{Yi Wang\textsuperscript{*}}, 
\textbf{Junxiao Liu\textsuperscript{*}}, 
\textbf{Shimao Zhang\textsuperscript{*}}, 
\textbf{Jiajun Chen}, 
\textbf{Shujian Huang\textsuperscript{\dag}} \\
National Key Laboratory for Novel Software Technology, Nanjing University \\
\texttt{\{yiw,junxiaoliu,smzhang\}@smail.nju.edu.cn} \\
\texttt{chenjj@nju.edu.cn, huangsj@nju.edu.cn}
}

\begin{document}
\maketitle

\renewcommand{\thefootnote}{\fnsymbol{footnote}}
\footnotetext[1]{Equal contribution.}
\footnotetext[2]{Corresponding author.}
\renewcommand{\thefootnote}{\arabic{footnote}}  % optional
\footnotetext[1]{The project is available at \url{https://github.com/NJUNLP/PATS}}

\begin{abstract}
Current large-language models (LLMs) typically adopt a fixed reasoning strategy, either simple or complex, for all questions, regardless of their difficulty. This neglect of variation in task and reasoning process complexity leads to an imbalance between performance and efficiency. Existing methods attempt to implement training-free fast-slow thinking system switching to handle problems of varying difficulty, but are limited by coarse-grained solution-level strategy adjustments. To address this issue, we propose a novel reasoning paradigm: \textit{Process-Level Adaptive Thinking Mode Switching (\textbf{PATS})}, which enables LLMs to dynamically adjust their reasoning strategy based on the difficulty of each step, optimizing the balance between accuracy and computational efficiency. Our approach integrates Process Reward Models (PRMs) with Beam Search, incorporating
progressive mode switching and bad-step penalty mechanisms. Experiments on diverse mathematical benchmarks demonstrate that our methodology achieves high accuracy while maintaining moderate token usage. This study emphasizes the significance of process-level, difficulty-aware reasoning strategy adaptation, offering valuable insights into efficient inference for LLMs.
\end{abstract}

\section{Introduction}

In recent years, Large Language Models (LLMs) have advanced from ``fast thinking'' to ``slow thinking'' by employing more sophisticated reasoning paradigms, leading to stronger performance on complex tasks. From direct answering to Chain-of-Thought (CoT) prompting~\cite{wei2022chain} and reflection mechanisms inspired by the o1-like system~\cite{jaech2024openai,guo2025deepseek}, LLMs have continued to enhance their reasoning capabilities. 

Although complex reasoning strategies like o1 perform well on challenging problems, they are inefficient when applied to simple ones~\cite{chen2024not}. In contrast, simple strategies such as direct answering excel at easy questions but struggle with complex ones. In practice, question difficulty varies, yet current models typically adopt a fixed reasoning strategy regardless of difficulty, leading to an imbalance between performance and computational efficiency. Meanwhile, research has shown that in mathematical reasoning, sub-steps requiring intensive computation pose the primary challenge for limited-scale supervised fine-tuned models~\cite{sun2025climbing}. Similarly, in reasoning tasks such as maze navigation, subproblems with varying levels of difficulty exist~\cite{saha2024system}. These findings indicate that the difficulty in the reasoning process changes dynamically, necessitating the dynamic allocation of more resources to harder sub-steps and adaptive adjustment of the reasoning strategy accordingly. Fixed strategies fail to account for variations in problem and process difficulty, highlighting the necessity of Adaptive adjustments to balance both accuracy and efficiency.

Inspired by the dual-process theory~\cite{wason1974dual,kahneman2011thinking}, humans can flexibly switch between fast (System 1) and slow (System 2) thinking: the former is fast and efficient for simple tasks, while the latter is slow and deliberate, suited for complex problems. These two cognitive styles align closely with the varying reasoning strategies employed by LLMs. Prior work has explored mechanisms for switching between System 1 and System 2 in LLMs, broadly categorized into training-based~\cite{su2024dualformer,saha2024system,cheng2025think} and training-free methods~\cite{yao2024hdflow}. This study focuses on the training-free setting. HDFLOW~\cite{yao2024hdflow} fixedly employs System 1 to generate an initial solution, and if the solution fails evaluation, the more complex System 2 is then activated to reconsider the problem. Deciding whether to switch thinking strategies only after obtaining a complete solution is overly coarse, misaligning with the current context of stepwise reasoning for complex problems and lacking adaptability to variations in the difficulty of the reasoning process.

To address these limitations, we propose Process-Level Adaptive Thinking Mode Switching (\textbf{PATS}), a novel reasoning paradigm that dynamically selects appropriate thinking mode at each reasoning step based on its difficulty, achieving a strong balance between accuracy and efficiency. Our approach is built on the following main designs: (1) We differentiate thinking modes based on the number of candidate steps generated at each step in Beam Search; (2) We assess the difficulty of each step using the PRM score for the top candidate, and adjust the reasoning strategy accordingly; (3) We incorporate mechanisms such as progressive switching, rapid error recovery, and appropriate rollback for bad steps. Compared to previous work, we performs finer-grained and more effective mode adjustments at the process-step level.

We conduct experiments across various policy models and PRMs, evaluating on multiple math reasoning benchmarks with varying difficulty levels. Experiments demonstrate that our approach consistently maintains high accuracy with moderate token usage, achieving an excellent balance between performance and efficiency. Furthermore, analysis of initial mode performance and the distribution of thinking modes across tasks of varying complexity further enhances our understanding of how reasoning strategies can be aligned with task difficulty.

Our main contributions are as follows:
\begin{itemize}[itemsep=0.5pt, topsep=2pt, parsep=0pt, partopsep=0pt]
  \item We propose a novel reasoning paradigm that adaptively switches between thinking modes based on the difficulty of each reasoning step, enabling real-time, step-level strategy adjustment.
  \item Our approach achieves a strong balance between performance and efficiency, consistently delivering high accuracy with moderate token usage, and providing insights into efficient reasoning with LLMs.
  \item Our analysis further highlights the necessity of more complex reasoning strategies as reasoning difficulty increases.
\end{itemize}

\section{Related Work}

\subsection{System Switching}

The dual-process theory~\cite{wason1974dual,kahneman2011thinking} posits that humans can switch between fast, intuitive System 1 and slow, deliberative System 2 to handle tasks of varying complexity. Inspired by this, many works introduce system switching into LLM reasoning, falling into two categories: training-based~\cite{su2024dualformer,saha2024system,cheng2025think} and training-free approaches~\cite{yao2024hdflow}. In this work, we focus on training-free approaches. HDFLOW~\cite{yao2024hdflow} adopts a fixed strategy where System 1 is first used to generate a complete solution. If the solution evaluation fails, System 2 is then invoked to re-solve the problem. The switch between thinking systems occurs after a complete solution is derived, representing a solution-level strategy adjustment with coarse granularity. In contrast, our approach enables finer-grained, step-level mode switching during the reasoning process.

\subsection{Test-time Scaling}

Test-time Scaling improves model performance by increasing inference computation cost~\cite{jaech2024openai,openai2024reasoning}, typically in parallel or sequential forms. Representative parallel approaches include Best-of-N decoding at the solution level~\cite{lightman2023let} and Beam Search at the step level~\cite{snell2024scaling}, the latter being a guided-search method~\cite{wang2023math,zhang2025process,she2025r} relying on Process Reward Models (PRMs) to guide step selection. PRMs offer step-level correctness estimation and have shown superior performance~\cite{uesato2022solving,lightman2023let,li202512surveyreasoning}. Beam Search benefits from larger candidate sets per step but incurs higher computation costs~\cite{wang2025don}. We adopt a PRM-guided Beam Search framework that enables dynamic switching between thinking modes.

\begin{figure*}[t]
    \centering
\includegraphics[width=0.9\textwidth]
{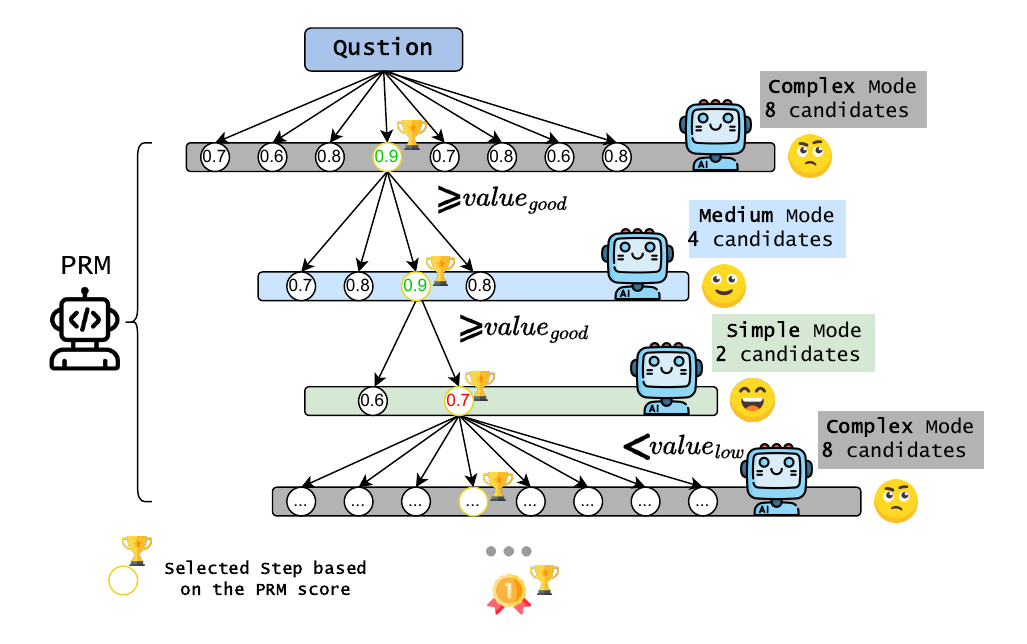}
    \caption{\textbf{Illustration of our Process-Level Adaptive Thinking Mode Switching (PATS) paradigm}. For math reasoning tasks, the model performs PRM-guided beam search, where the number of candidate steps (2/4/8) serves as a proxy for thinking mode complexity. At each step, the policy model selects the top-scoring candidate , infers reasoning difficulty, and dynamically switches to the corresponding mode. The framework further incorporates progressive switching, rapid error recovery, and appropriate penalization mechanisms.}
    \label{fig:pipeline}
\end{figure*}

\subsection{Reasoning Process}

Reasoning tasks have long been a central focus for LLMs. Studies show that some steps in hard-level problems are computation-intensive, posing challenges for models with limited-scale SFT~\cite{sun2025climbing}. In tasks like maze navigation, subproblems vary in difficulty and require adaptive allocation of computational resources~\cite{saha2024system}. These findings indicate that reasoning difficulty changes dynamically throughout the reasoning process, and thus the reasoning strategy should adapt accordingly. We emphasize a process-level, difficulty-aware adaptive reasoning strategy to achieve a balance between accuracy and efficiency.

\section{Method}
\label{sec:method}

In this section, we propose the Process-Level Adaptive Thinking Mode Switching (\textbf{PATS}) pipeline. First, in \S\ref{sec:prelim}, we review the beam search guided by Process Reward Model (PRM) and describe our background setup. Then, in \S\ref{sec:pipeline}, we present the overall design of the PATS pipeline. Finally, in \S\ref{sec:switching}, we elaborate on the core component: the thinking mode switching mechanism. The full pipeline is illustrated in Figure~\ref{fig:pipeline}.

\subsection{Preliminary Study}
\label{sec:prelim}

PRM-guided beam search is a representative test-time scaling method~\cite{snell2024scaling}. It scores candidate steps generated by the policy model at each step and selects the top-$k$ candidates with the highest scores for the next iteration. This paper focuses on the best-first search (BFS) scenario with $K=1$ as the reasoning context~\citep{wang2025don}.

Given a problem $q$, the policy model performs step-by-step reasoning, generating a solution path $S = \{s_1, s_2, ..., s_n\}$, where $s_n$ denotes the $n$-th reasoning step. BFS consists of two iterative operations: expansion and selection.

In the expansion stage of step $i$, the model generates $w$ candidate steps:

\begin{equation}
C_i = \{c_{i,1}, c_{i,2}, ..., c_{i,w}\}
\end{equation}

Here, $C_i$ denotes all candidate steps at step $i$, $w$ is a hyperparameter controlling search width, and $c_{i,j}$ is the $j$-th candidate step at step $i$.

In the selection phase of step $i$, BFS chooses the candidate step with the highest PRM score $v(\cdot)$ from $C_i$ as the final selected step $s_i$ , proceeding to the next iteration:
\begin{equation}
s_i = \arg\max_{c \in C_i} v(c)
\end{equation}
This iteration continues until the final answer is generated. Through iterative expansion and optimal selection, BFS efficiently attains a final solution.

PRM-guided beam search uses a fixed width $w$ at every step. Prior studies~\cite{wang2025don} show that increasing $w$ improves performance but significantly raises computational cost. This trade-off, where increased computational cost leads to improved performance, resembles a model adopting more complex reasoning modes to solve more challenging problems by expending additional computational resources. Building on this foundation, we use PRM-guided beam search framework to simulate problem-solving scenarios across a range of complexity levels.

\subsection{Process-Level Adaptive Thinking Mode Switching}
\label{sec:pipeline}

As discussed in \S\ref{sec:prelim}, a larger search width $w$ results in higher accuracy but greater computation. Current PRM-guided beam search typically employs a predefined, fixed $w$ at each step, implying a unchanged reasoning strategy throughout the inference process. Therefore, at each step, a larger search width $w$ can be considered to represent a more complex thinking mode.

Let $w_i$ denote the number of candidates generated at step $i$, and $mode_i$ represent the reasoning mode employed at that step. We define the reasoning strategy at step $i$ based on the number of candidates $w_i$ as follows:

\begin{itemize}
    \item \textbf{Simple Thinking Mode}: \( mode_i = \text{simple} \), corresponding to \( w_i = 2 \)
    \item \textbf{Medium Thinking Mode}: \( mode_i = \text{medium} \), corresponding to \( w_i = 4 \)
    \item \textbf{Complex Thinking Mode}: \( mode_i = \text{complex} \), corresponding to \( w_i = 8 \)
\end{itemize}

Considering that mathematical reasoning is characterized by step-by-step execution and generalizability, we choose it as our task. The proposed Process-Level Adaptive Thinking Mode Switching (\textbf{PATS}) framework works as follows:

The policy model performs step-by-step reasoning under PRM-guided beam search, selecting the candidate step with the highest PRM score at each step. Based on this score, the model assesses the current reasoning difficulty and adaptively switches to an appropriate thinking mode. By default, the model starts reasoning in complex mode ($mode_1 = complex$). 

\subsection{Thinking Mode Switching Mechanism}
\label{sec:switching}

The core of PATS is to dynamically adjust the thinking mode based on the reasoning difficulty of the current state. As the reasoning state becomes more difficult, a correspondingly more complex reasoning mode is required. A higher PRM score for the current step indicates better reasoning quality, suggesting that the model is in a more favorable reasoning state and is more likely to be on the correct reasoning path~\cite{wang2023math}. Studies use PRM score of the final answer to evaluate problem difficulty~\citep{snell2024scaling}, with higher scores indicating easier problems for the model. Additionally, inspired by reward signal can guide resource allocation~\citep{sun2024fast}, \citet{fu2024efficiently} collect the terminal reward scores from each reasoning path and aggregate them, where a higher aggregated reward indicates that less computational resources need to be allocated.

Thus, the PRM score can acts as an indicator of current reasoning difficulty. We hypothesize that a higher PRM score at the current step indicates a lower reasoning difficulty for that step, allowing for the allocation of fewer computational resources and the adoption of simpler reasoning strategies. Since adjacent steps are likely to share similar states, if the current step is easy to reason, suggesting that the next step is also more likely to be easy and thus allows switching to a simpler mode. 

After selecting the final step $s_i$ at step $i$, we use $v(s_i)$, the PRM score of the chosen step, to estimate the current reasoning difficulty and adapt the reasoning strategy for the subsequent step accordingly.

\begin{itemize}
    \item If $v(s_i) \geq \text{value}_{\text{good}}$, the reasoning state is considered favorable, and the thinking mode should switch to a simpler one. Inspired by the approach of employing progressively shorter reasoning trajectories during training~\citep{ma2025cot}, we implement a smooth mode transition when the reasoning state is favorable. Specifically, if the current step $i$ is in complex thinking mode ($mode_i = complex$), the next step becomes medium thinking mode ($mode_{i+1} = medium$). Similarly, if $mode_i = medium$, then $mode_{i+1} = simple$; if $mode_i = simple$, then $mode_{i+1} = simple$.
    \item If $v(s_i) < \text{value}_{\text{low}}$, the reasoning state is considered unfavorable, and the thinking mode should switched to the most complex setting to mitigate potential error accumulation. This means that, regardless of the current mode $mode_i$, the reasoning mode for the next step is set to $mode_{i+1} = complex$.
    \item If $v(s_i)$ does not satisfy the threshold conditions for mode switching, the current thinking mode is retained for the next step; that is, $mode_{i+1} = mode_i$.
\end{itemize}

In addition to adapting the reasoning strategy for the subsequent step, this mechanism includes immediate strategy adjustments for critically poor steps. If $v(s_i)$ falls significantly below $\text{value}_{\text{low}}$, it suggests a severe error at the current step, and delaying correction until the next step may be too late to prevent further reasoning failure. Thus, we define a critical threshold $\text{value}_{\text{bad}}$. If $v(s_i) < \text{value}_{\text{bad}}$, we penalize the current step and restart step $i$ in complex mode by regenerating $w_i = 8$ candidates. To avoid infinite loops on unresolvable steps, each step is penalized at most once.

\section{Experiment}

\subsection{Experimental Setup}

To assess the effectiveness of our approach, we conduct experiments on policy models with different parameter scales and a variety of PRMs, evaluated on multiple mathematical reasoning benchmarks.

\paragraph{Datasets.}
We evaluate on widely used mathematical reasoning benchmarks, including \textbf{GSM8k}~\cite{cobbe2021training}, \textbf{MATH500}~\cite{lightman2023let}, \textbf{Minerva Math}~\cite{lewkowycz2022solving}, \textbf{AMC23}, and \textbf{AIME24}, which collectively span elementary, intermediate, and advanced levels of mathematical reasoning.

\paragraph{Policy Models.}
We utilize the Instruct variants of the \textbf{Qwen2.5} family~\cite{yang2024qwen2} and conduct experiments with models of varying parameter sizes. Our primary experiments are conducted using the \textbf{Qwen2.5-7B-Instruct} model.

\paragraph{Process Reward Models (PRMs).}
Our evaluation includes a selection of open-source PRMs. Unless otherwise specified, the main results are reported using \textbf{Qwen2.5-Math-PRM-7B}~\cite{yang2024qwen2math}.

\begin{itemize}
    \item \textbf{Math-Shepherd}~\cite{wang2023math}: Uses Monte Carlo Tree Search (MCTS) to estimate the probability of reaching the correct solution as step-level labels.
    \item \textbf{Qwen2.5-Math-PRM-7B}~\cite{yang2024qwen2math}: Based on Math-Shepherd, it further incorporates LLM-as-Judge to perform consistency filtering.
    \item \textbf{Qwen2.5-Math-7B-PRM800K}~\cite{zheng2024processbench}: Obtained by fine-tuning Qwen2.5-Math-7B-Instruct on the PRM800K dataset.
\end{itemize}

\paragraph{Evaluation Metrics.}
We evaluate model performance along two dimensions: \textbf{accuracy} and \textbf{efficiency}. For accuracy, we report the mean answer correctness. For efficiency, we follow~\cite{kang2024mindstar,wang2024litesearch} and compute the average number of generated output tokens.

\paragraph{Hyperparameters and Thresholds.}
We set the temperature to 0.6 to balance candidate step diversity and generation quality~\cite{zhang2024edt}. We set reward thresholds as follows: $value_{good} = 0.85$, $value_{low} = 0.75$, and $value_{bad} = 0.4$. The threshold setting is determined by the distribution of scoring preferences in the PRM and empirical configuration.

\subsection{Baselines}

We compared fixed and random thinking modes with Beam Search to assess the effectiveness of adaptive mode switching. Furthermore, we contrast our fine-grained control strategy with existing coarse-grained approaches to validate the advantages of more precise adjustments. The evaluated baselines are as follows:

\begin{itemize}

    \item \textbf{All-simple}: It uses a fixed \textit{simple} mode at every step, generating two candidate steps per iteration. Formally, this corresponds to $mode_1 = mode_2 = \ldots = mode_n = simple$, resembling direct-answer-style inference and suitable for relatively simple problems.
    
    \item \textbf{All-medium}: It follows the same structure as All-simple, but adopts a fixed \textit{medium} mode at each step, generating four candidate steps per iteration. This reflects typical chain-of-thought reasoning, suited for intermediate problems.
    
    \item \textbf{All-complex}: It continues this fixed pattern by using the \textit{complex} mode throughout, with eight candidate steps per iteration. This reflects o1-style slow thinking and is better suited for more challenging problems.
    
    \item \textbf{Random-mode Switch}: It begins with the \textit{complex} thinking mode and, at each subsequent step, randomly selects one of the three modes (\textit{simple}, \textit{medium}, or \textit{complex}) with equal probability.
    
\begin{table*}[!t]
\centering
\scriptsize
\setlength{\tabcolsep}{4pt}
\renewcommand{\arraystretch}{1.15}
\begin{tabular}{>{\bfseries}l 
                c c c c c c c c c c 
                >{\columncolor{gray!20}}c 
                >{\columncolor{gray!20}}c}
\toprule
\textbf{Setting} 
& \multicolumn{2}{c}{\textbf{GSM8K}} 
& \multicolumn{2}{c}{\textbf{MATH500}} 
& \multicolumn{2}{c}{\textbf{AMC23}} 
& \multicolumn{2}{c}{\textbf{MinervaMATH}} 
& \multicolumn{2}{c}{\textbf{AIME24}} 
& \multicolumn{2}{>{\columncolor{gray!20}}c}{\textbf{Average}} \\
\cmidrule(r){2-3} \cmidrule(r){4-5} \cmidrule(r){6-7} 
\cmidrule(r){8-9} \cmidrule(r){10-11} \cmidrule(r){12-13}
& Acc~$\uparrow$ & Token~$\downarrow$ 
& Acc~$\uparrow$ & Token~$\downarrow$ 
& Acc~$\uparrow$ & Token~$\downarrow$ 
& Acc~$\uparrow$ & Token~$\downarrow$ 
& Acc~$\uparrow$ & Token~$\downarrow$ 
& Acc~$\uparrow$ & Token~$\downarrow$ \\
\midrule
All-simple & 93.8 & 564.8 & 76.2 & 1154.3 & 52.5 & 1742.5 & 38.2 & 1234.3 & 16.7 & 2075.7 
& 55.5 & 1354.3 \\
All-medium & 94.4 & 1001.0 & 80.2 & 2204.2 & 57.5 & 3337.3 & 43.0 & 2431.6 & 16.7 & 4113.3 
& 58.4 & 2617.5 \\
All-complex & 94.9 & 1774.9 & 81.0 & 4068.9 & 67.5 & 6231.6 & 44.5 & 4571.9 & 20.0 & 8705.6 
& 61.6 & 5070.6 \\
Solution-verification Switch & 94.5 & 596.3 & 79.6 & 2110.8 & 55.0 & 4318.6 & 38.6 & 1466.7 & 16.7 & 6621.1 
& 56.9 & 3022.7 \\
Random-mode Switch & 94.3 & 1214.7 & 79.0 & 2838.1 & 52.5 & 4858.9 & 43.0 & 2996.2 & 13.3 & 6330.0 
& 56.4 & 3647.6 \\
\ding{72} PATS (Ours) & 94.8 & 855.8 & 80.6 & 2067.7 & 65.0 & 3365.7 & 43.0 & 1929.9 & 23.3 & 5821.0 
& 61.3 & 2808.0 \\
\bottomrule
\end{tabular}
\caption{Comparison of average accuracy and token usage on five math reasoning benchmarks. Based on the average metrics, Our method PATS achieves effective and efficient reasoning with an excellent accuracy-efficiency balance.}
\label{tab:main-results}
\end{table*}

    \item \textbf{Solution-verification Switch}: It first generates a complete solution using the \textit{All-simple} strategy and uses the PRM score at the final step as the solution-level score, following \cite{snell2024scaling}. If the score is greater than or equal to the threshold $value_{low}$, the solution is accepted; otherwise, we retry using the \textit{All-medium} strategy. If this fails again, we switch to the \textit{All-complex} strategy for final reasoning and answer generation. This approach simulates coarse-grained solution-level switching: fast reasoning is first used to generate a complete solution, and more complex strategies are applied only if it fails verification.
\end{itemize}

\begin{table*}[t]
\centering
\scriptsize
\renewcommand{\arraystretch}{1.15}
\begin{tabular}{l c c c c c c c c c c}
\toprule
\multirow{2}{*}{\textbf{Setting}} 
& \multicolumn{4}{c}{\textbf{Level 1 (Easy)}} 
& \multicolumn{4}{c}{\textbf{Level 2 (Medium)}} 
& \multicolumn{2}{c}{\textbf{Level 3 (Hard)}} \\
\cmidrule(r){2-5} \cmidrule(r){6-9} \cmidrule(r){10-11}
& \multicolumn{2}{c}{\textbf{GSM8K}} 
& \multicolumn{2}{c}{\textbf{MATH500}} 
& \multicolumn{2}{c}{\textbf{AMC23}} 
& \multicolumn{2}{c}{\textbf{MinervaMATH}} 
& \multicolumn{2}{c}{\textbf{AIME24}} \\
\cmidrule(r){2-3} \cmidrule(r){4-5} \cmidrule(r){6-7} \cmidrule(r){8-9} \cmidrule(r){10-11}
& Acc~$\uparrow$ & Token~$\downarrow$ 
& Acc~$\uparrow$ & Token~$\downarrow$ 
& Acc~$\uparrow$ & Token~$\downarrow$ 
& Acc~$\uparrow$ & Token~$\downarrow$ 
& Acc~$\uparrow$ & Token~$\downarrow$ \\
\midrule
\textbf{PATS-first-simple} & 94.4 & 600.1 & 80.2 & 1582.0 & 62.5 & 2400.1 & 43.0 & 1648.4 & 16.7 & 3998.5 \\
\textbf{PATS-first-medium} & 94.7 & 644.1 & 80.2 & 1687.8 & 65.0 & 3349.2 & 44.1 & 1633.9 & 16.7 & 5776.0 \\
\textbf{\ding{72}~PATS (Ours)} & 94.8 & 855.8 & 80.6 & 2067.7 & 65.0 & 3365.7 & 43.0 & 1929.9 & 23.3 & 5821.0 \\
\bottomrule
\end{tabular}
\caption{Performance of different initial thinking modes across tasks of varying difficulty. Results show that aligning the initial thinking mode with task difficulty leads to better performance.}
\label{tab:mode-init}
\end{table*}

Among the fixed-mode baselines, performance consistently ranks as: \textbf{All-complex > All-medium > All-simple} in terms of both average accuracy and output tokens.

\section{Results}

\subsection{Main Results}

We report the performance of various baselines and our adaptive thinking mode switching method in Table~\ref{tab:main-results}, comparing their performance in terms of average accuracy and average output token count. Our findings are as follows:

\textbf{PATS achieves both high accuracy and efficiency.} 
 Across tasks of varying difficulty, PATS achieves an average accuracy close to that of the All-complex setting (only 0.3 points lower), while using just 55.4\% of its output tokens. Compared to All-medium, it achieves nearly 3 points higher accuracy with comparable token usage. Relative to All-simple, it yields a substantial 5.8-point improvement in accuracy. These results demonstrate that PATS maintains high average accuracy while achieving low token usage.

\textbf{PATS achieves a strong accuracy-efficiency balance.} As shown in Table~\ref{tab:main-results}, when token usage doubles, PATS achieves a significant improvement in accuracy over All-simple---twice the gain achieved by All-medium under similar resource consumption, demonstrating higher resource efficiency. While All-complex requires nearly twice the resources of All-medium to achieve comparable accuracy gains, our PATS achieves similar improvements with only a 7.3\% increase in token usage, thereby significantly reducing resource consumption. These results highlight PATS's superior balance of accuracy and efficiency.

\textbf{PATS outperforms coarse-grained switching.} As shown in Table~\ref{tab:main-results}, PATS surpasses Solution-verification Switch by 4.4 points in average accuracy while using about 7\% fewer tokens. These results demonstrate that our fine-grained, process-level reasoning strategy adjustment outperforms coarse-grained, solution-level switching, underscoring the importance of timely strategy adaptation throughout the reasoning process. Delaying the switch until a complete solution is generated fails to adapt to the varying difficulty of the reasoning process.

\begin{figure*}[htbp]
  \centering
  \includegraphics[width=\linewidth]{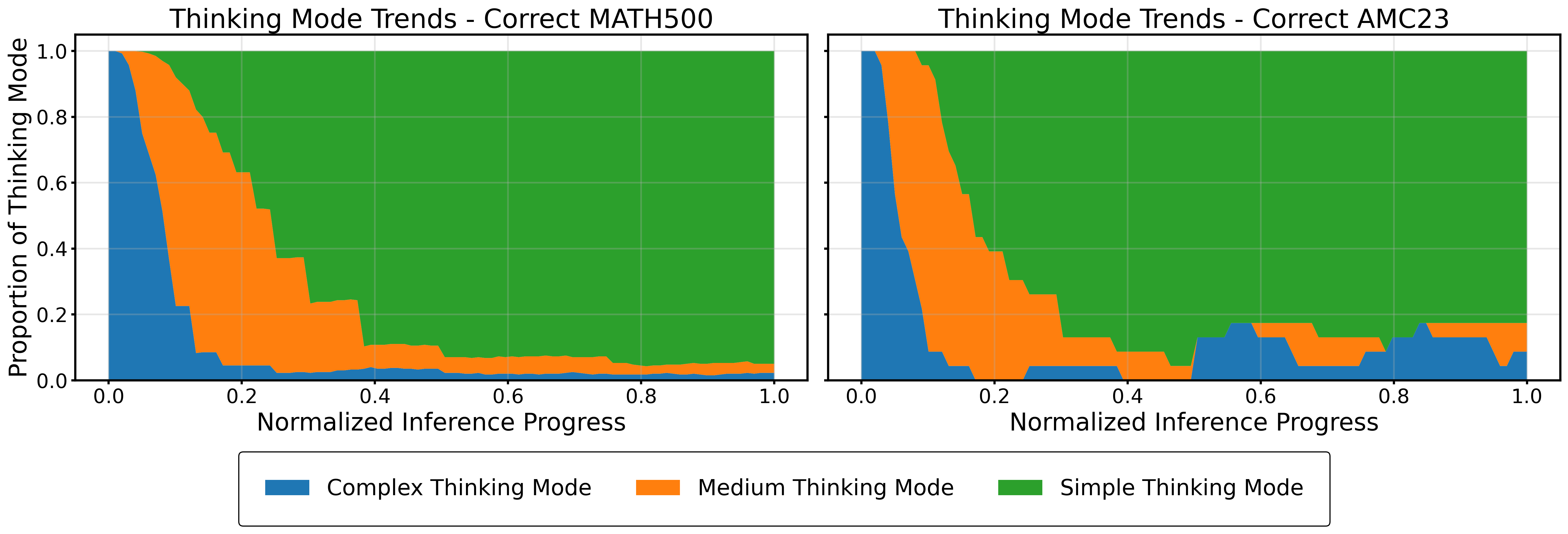}
  \caption{Comparison of thinking mode distributions over the reasoning process for MATH500 (easier) and AMC23 (harder) tasks. AMC23 shows a noticeably higher proportion of more complex thinking modes in the mid-to-late stages, indicating greater reasoning effort consistent with higher task difficulty.}
  \label{fig:math500-trend}
\end{figure*}

\begin{table*}[t]
\centering
\scriptsize
\setlength{\tabcolsep}{4pt}
\renewcommand{\arraystretch}{1.15}
\begin{tabular}{
    >{\bfseries}l 
    c c c c c c c c c c 
    >{\columncolor{gray!20}}c 
    >{\columncolor{gray!20}}c}
\toprule
\textbf{Setting} 
& \multicolumn{2}{c}{\textbf{GSM8K}} 
& \multicolumn{2}{c}{\textbf{MATH500}} 
& \multicolumn{2}{c}{\textbf{AMC23}} 
& \multicolumn{2}{c}{\textbf{MinervaMATH}} 
& \multicolumn{2}{c}{\textbf{AIME24}} 
& \multicolumn{2}{>{\columncolor{gray!20}}c}{\textbf{Average}} \\
\cmidrule(r){2-3} \cmidrule(r){4-5} \cmidrule(r){6-7} \cmidrule(r){8-9} \cmidrule(r){10-11} \cmidrule(r){12-13}
& Acc~$\uparrow$ & Token~$\downarrow$ 
& Acc~$\uparrow$ & Token~$\downarrow$ 
& Acc~$\uparrow$ & Token~$\downarrow$ 
& Acc~$\uparrow$ & Token~$\downarrow$ 
& Acc~$\uparrow$ & Token~$\downarrow$ 
& Acc~$\uparrow$ & Token~$\downarrow$ \\
\midrule
\textbf{PATS-No-Penalty} & 93.9 & 829.4 & 79.0 & 1931.3 & 47.5 & 2604.0 & 42.6 & 1836.8 & 16.7 & 4612.9 & 55.9 & 2362.9 \\
\textbf{PATS-Infinite-Penalty} & 95.0 & 920.3 & 81.4 & 2307.8 & 60.0 & 3880.9 & 43.8 & 2188.5 & 20.0 & 11621.5 & 60.0 & 4183.8 \\
\textbf{\ding{72} PATS (Ours)} & 94.8 & 855.8 & 80.6 & 2067.7 & 65.0 & 3365.7 & 43.0 & 1929.9 & 23.3 & 5821.0 & 61.3 & 2808.0 \\
\bottomrule
\end{tabular}
\caption{Effect of penalizing bad reasoning steps in adaptive switching. Based on the average metrics, PATS achieves the best balance—significantly improving accuracy while avoiding the high token cost on unresolvable steps.}
\label{tab:penalty-study}
\end{table*}

\textbf{PATS's pipeline is rational and effective.} PATS not only outperforms the Random-mode Switch by 4.9 points in average accuracy, but also reduces average token consumption by 23\%, as shown in Table~\ref{tab:main-results}. Under the same process-level switching framework, PATS significantly outperforms random switching across all dimensions. Notably, random switching yields performance comparable to coarse-grained solution-level switching, suggesting that random mode integration alone fails to fully exploit the advantages of process-level adaptability. This underscores the rationality and effectiveness of our pipeline's switching strategy design.

In summary, PATS dynamically adjusts reasoning strategies based on step-wise difficulty, has been empirically validated. The results demonstrate that well-designed adaptive thinking mode switching, by rationally allocating computational resources during the inference process, achieves an excellent accuracy-efficiency balance while enabling effective and efficient reasoning.

\subsection{Analysis and Generalization Experiments}
\label{sec:analysis-generalization}

This subsection provides a further analysis of the adaptive thinking mode switching process and conducts generalization experiments on both the policy model and PRM.

\begin{table}[t]
\centering
\scriptsize
\setlength{\tabcolsep}{4pt}
\renewcommand{\arraystretch}{1.15}
\begin{tabular}{l c c}
\toprule
\textbf{Setting} & \textbf{Average Accuracy~$\uparrow$} & \textbf{Average Token~$\downarrow$} \\
\midrule\midrule
\multicolumn{3}{l}{\textbf{Qwen2.5-1.5B-Instruct}} \\
\cmidrule(r){1-3}
All-simple & 36.3 & 1548.5 \\
All-medium & 40.1 & 3146.4 \\
All-complex & 46.7 & 6377.7 \\
\ding{72}~PATS (Ours) & 42.6 & 3994.1 \\
\midrule\midrule 
\multicolumn{3}{l}{\textbf{Qwen2.5-3B-Instruct}} \\
\cmidrule(r){1-3}
All-simple & 49.7 & 1437.2 \\
All-medium & 51.2 & 2995.9 \\
All-complex & 54.5 & 6126.7 \\
\ding{72}~PATS (Ours) & 53.0 & 3288.5 \\
\bottomrule
\end{tabular}
\caption{Generalization of PATS across different policy models, while keeping the Process Reward Model fixed as Qwen2.5-Math-PRM-7B.}
\label{tab:strategy-gen}
\end{table}

\paragraph{Performance of Initial Thinking Modes Across Varying Task Difficulties.} We investigate the impact of initiating reasoning with different thinking modes on model performance across tasks of varying difficulty. Based on greedy decoding accuracy, we categorize datasets into three levels of difficulty: easy (Level 1), medium (Level 2), and hard (Level 3). PATS defaults to initiating the first step with the complex thinking mode. Building on this design, we further investigate the effects of starting with simple and medium thinking modes. As shown in Table~\ref{tab:mode-init}, we analyze the trade-off between accuracy and token efficiency.
The results indicate that more difficult tasks derive greater benefit from initiating reasoning with more complex thinking modes. For easy tasks, initiating with the simple mode achieves the lowest token usage with comparable accuracy. For medium tasks, the medium mode achieves the highest accuracy with moderate token usage. For hard tasks, the complex mode clearly outperforms the others in terms of accuracy. These findings suggest that aligning the initial thinking mode with task difficulty effectively balances accuracy and computational efficiency.

\paragraph{Reasoning Behavior across Task Difficulties.}

We compare the reasoning processes across tasks of varying difficulty levels, selecting MATH500 as a representative of easier tasks and AMC23 for harder ones. To enable a meaningful comparison of reasoning trajectories across problems with differing numbers of steps, we normalize the reasoning process to the range \([0, 1]\) and partition it into five equal stages (e.g., 0--0.2 as the early stage, and so forth).

As shown in Figure~\ref{fig:math500-trend}, more complex modes indicate greater reasoning difficulty and effort used. For equally correct solutions, AMC23 exhibits a higher proportion of complex thinking modes during the mid-to-late stages of reasoning compared to MATH500, reflecting increased cognitive effort and aligning with the greater complexity of AMC23 problems.  This shows that harder tasks require more reasoning effort to achieve correct solutions, underscoring that greater reasoning difficulty necessitates more computational resources to ensure final answer correctness.

\paragraph{Necessity and Moderation of Penalty on Bad Steps.}

In PATS, steps with PRM scores below a threshold $value_{bad}$ are classified as bad steps. To prevent error propagation, we apply a one-time penalty by rethinking these steps with the complex mode. Table~\ref{tab:penalty-study} compares three strategies: No Penalty (bad steps are left unprocessed), Infinite Penalty (bad steps are rethought repeatedly in complex mode until the score exceeds the threshold), and PATS (our proposed one-time penalty method).

The results indicate that PATS strikes the best balance, achieving the highest accuracy(+5.4 over No Penalty and +1.3 over Infinite Penalty)while maintaining significantly lower token usage than Infinite Penalty and comparable usage to No Penalty. This demonstrates the importance of penalizing suboptimal reasoning steps to prevent delayed correction, while also underscoring the need for moderation to avoid excessive rethinking on unresolvable steps, which would otherwise lead to unnecessary token consumption. By applying a one-time penalty, PATS limits error propagation without incurring extra computational cost, highlighting the value of controlled intervention in reasoning.

\begin{table}[t]
\centering
\scriptsize
\setlength{\tabcolsep}{4pt}
\renewcommand{\arraystretch}{1.15}
\begin{tabular}{l c c}
\toprule
\textbf{Setting} & \textbf{Average Accuracy~$\uparrow$} & \textbf{Average Token~$\downarrow$} \\
\midrule\midrule
\multicolumn{3}{l}{\textbf{Math-Shpherd}} \\
\cmidrule(r){1-3}
All-simple & 53.5 & 1286.8 \\
All-medium & 54.1 & 2270.8 \\
All-complex & 54.5 & 4167.8 \\
\ding{72}~PATS (Ours) & \textbf{55.5} & 2732.1 \\
\midrule\midrule
\multicolumn{3}{l}{\textbf{Qwen2.5-Math-7B-PRM800K}} \\
\cmidrule(r){1-3}
All-simple & 55.4 & 1356.5 \\
All-medium & 56.3 & 2743.4 \\
All-complex & 58.3 & 5275.9 \\
\ding{72}~PATS (Ours) & 57.5 & 2418.0 \\
\bottomrule
\end{tabular}
\caption{Generalization of PATS across different Process Reward Models, while keeping the policy model fixed as Qwen2.5-7B-Instruct.}
\label{tab:prm-gen}
\end{table}

\paragraph{Generalization of Policy and Process Reward Models.}

We primarily examine the generalization capabilities of two key components in PATS: the policy model and the process reward model.  The average results are reported in Table~\ref{tab:strategy-gen} and Table~\ref{tab:prm-gen} (complete results are provided in Appendix~\ref{sec:appendix}).

As shown in Table~\ref{tab:strategy-gen} and Table~\ref{tab:prm-gen} , the proposed pipeline exhibits strong generalization across varying policy model scales and PRMs. In all three scenarios, PATS consistently outperforms All-simple and All-medium in average accuracy, while maintaining moderate token usage close to All-medium. Notably, in the Math-Shepherd setting, our method even outperforms All-complex in terms of accuracy. These results highlight the robustness of our adaptive paradigm across a wide range of policy models and PRMs.

\section{Conclusion}

In this paper, we propose a novel reasoning paradigm—\textbf{Process-Level Adaptive Thinking Mode Switching (PATS)}. This method leverages PRM scores during the reasoning process to estimate the current difficulty and accordingly adjusts the appropriate reasoning mode, enabling rational and dynamic allocation of computational resources. Experiments show it achieves high accuracy with low computational cost across multiple math reasoning datasets. This paradigm highlights that fine-grained, real-time adjustment of reasoning strategies based on process-level difficulty can effectively balance accuracy and efficiency, offering new insights into efficient reasoning with LLMs.

\section{Limitations}

Due to computational constraints, our experiments were limited to relatively small-scale policy models (1.5B, 3B, and 7B), and have not yet been validated on larger models. Extending our experiments to larger-scale models could further deepen the understanding of our proposed paradigm. Additionally, our approach relies on a process reward model as a key evaluation component. Incorporating alternative evaluation methods—such as LLM-as-Judge or generative reward models—could expand the scope of our experiments. We look forward to exploring these directions in future work.

% \section*{Acknowledgments}

% This document has been adapted
% by Steven Bethard, Ryan Cotterell and Rui Yan
% from the instructions for earlier ACL and NAACL proceedings, including those for
% ACL 2019 by Douwe Kiela and Ivan Vuli\'{c},
% NAACL 2019 by Stephanie Lukin and Alla Roskovskaya,
% ACL 2018 by Shay Cohen, Kevin Gimpel, and Wei Lu,
% NAACL 2018 by Margaret Mitchell and Stephanie Lukin,
% Bib\TeX{} suggestions for (NA)ACL 2017/2018 from Jason Eisner,
% ACL 2017 by Dan Gildea and Min-Yen Kan,
% NAACL 2017 by Margaret Mitchell,
% ACL 2012 by Maggie Li and Michael White,
% ACL 2010 by Jing-Shin Chang and Philipp Koehn,
% ACL 2008 by Johanna D. Moore, Simone Teufel, James Allan, and Sadaoki Furui,
% ACL 2005 by Hwee Tou Ng and Kemal Oflazer,
% ACL 2002 by Eugene Charniak and Dekang Lin,
% and earlier ACL and EACL formats written by several people, including
% John Chen, Henry S. Thompson and Donald Walker.
% Additional elements were taken from the formatting instructions of the \emph{International Joint Conference on Artificial Intelligence} and the \emph{Conference on Computer Vision and Pattern Recognition}.


\newpage

\appendix

\section{The complete results of the generalization experiments}
\label{sec:appendix}

We report the complete results of generalization experiments conducted on policy models of varying scales and different process reward models. The complete results for the policy model generalization experiments are presented in Table~\ref{tab:strategy-generalization}, and the complete results for the process reward model generalization experiments are presented in Table~\ref{tab:prm-generalization}.

\begin{table*}[t]
\centering
\scriptsize
\setlength{\tabcolsep}{4pt}
\renewcommand{\arraystretch}{1.15}
\begin{tabular}{l c c c c c c c c c c >{\columncolor{gray!20}}c >{\columncolor{gray!20}}c}
\toprule
\textbf{Setting} 
& \multicolumn{2}{c}{\textbf{GSM8K}} 
& \multicolumn{2}{c}{\textbf{MATH500}} 
& \multicolumn{2}{c}{\textbf{AMC23}} 
& \multicolumn{2}{c}{\textbf{MinervaMATH}} 
& \multicolumn{2}{c}{\textbf{AIME24}} 
& \multicolumn{2}{c}{\cellcolor{gray!20}\textbf{Average}} \\
\cmidrule(r){2-3} \cmidrule(r){4-5} \cmidrule(r){6-7} \cmidrule(r){8-9} \cmidrule(r){10-11} \cmidrule(r){12-13}
& Acc~$\uparrow$ & Token~$\downarrow$ 
& Acc~$\uparrow$ & Token~$\downarrow$ 
& Acc~$\uparrow$ & Token~$\downarrow$ 
& Acc~$\uparrow$ & Token~$\downarrow$ 
& Acc~$\uparrow$ & Token~$\downarrow$ 
& \cellcolor{gray!20}Acc~$\uparrow$ & \cellcolor{gray!20}Token~$\downarrow$ \\
\midrule\midrule
\multicolumn{13}{l}{\textbf{Qwen2.5-1.5B-Instruct}} \\
\cmidrule(r){1-13}
All-simple & 69.0 & 574.9 & 57.0 & 1343.3 & 32.5 & 1874.0 & 16.5 & 1244.7 & 6.7 & 2705.7 & 36.3 & 1548.5 \\
All-medium & 79.8 & 1188.8 & 66.6 & 2667.2 & 30.0 & 4030.1 & 17.3 & 2647.8 & 6.7 & 5198.1 & 40.1 & 3146.4 \\
All-complex & 84.8 & 2345.3 & 69.4 & 5519.0 & 45.0 & 7772.2 & 24.3 & 5507.7 & 10.0 & 10744.3 & 46.7 & 6377.7 \\
\ding{72}~PATS (Ours) & 83.2 & 1387.5 & 66.8 & 3127.7 & 37.5 & 4757.9 & 22.4 & 3531.4 & 3.3 & 7166.1 & 42.6 & 3994.1 \\
\midrule\midrule
\multicolumn{13}{l}{\textbf{Qwen2.5-3B-Instruct}} \\
\cmidrule(r){1-13}
All-simple & 88.6 & 616.1 & 69.4 & 1235.1 & 50.0 & 1758.0 & 33.8 & 1431.5 & 6.7 & 2145.2 & 49.7 & 1437.2 \\
All-medium & 92.3 & 1185.2 & 74.4 & 2420.1 & 45.0 & 3609.1 & 37.5 & 2903.2 & 6.7 & 4862.0 & 51.2 & 2995.9 \\
All-complex & 93.7 & 2203.8 & 75.8 & 4819.7 & 52.5 & 6939.7 & 37.1 & 5951.5 & 13.3 & 10718.8 & 54.5 & 6126.7 \\
\ding{72}~PATS (Ours) & 92.7 & 1126.6 & 75.8 & 2376.2 & 55.0 & 4083.7 & 34.9 & 2969.2 & 6.7 & 5886.6 & 53.0 & 3288.5 \\
\bottomrule
\end{tabular}
\caption{All results from the policy model generalization experiments are presented. Each policy model is evaluated on five math benchmarks. \textbf{All-simple}, \textbf{All-medium}, and \textbf{All-complex} perform fixed-mode reasoning with 2/4/8 candidates per step respectively. \textbf{PATS (Ours)} dynamically switches among modes at each step according to the PRM scores of the intermediate steps.}
\label{tab:strategy-generalization}
\end{table*}

\begin{table*}[t]
\centering
\scriptsize
\setlength{\tabcolsep}{4pt}
\renewcommand{\arraystretch}{1.15}
\begin{tabular}{l c c c c c c c c c c >{\columncolor{gray!20}}c >{\columncolor{gray!20}}c}
\toprule
\textbf{Setting} 
& \multicolumn{2}{c}{\textbf{GSM8K}} 
& \multicolumn{2}{c}{\textbf{MATH500}} 
& \multicolumn{2}{c}{\textbf{AMC23}} 
& \multicolumn{2}{c}{\textbf{MinervaMATH}} 
& \multicolumn{2}{c}{\textbf{AIME24}} 
& \multicolumn{2}{c}{\cellcolor{gray!20}\textbf{Average}} \\
\cmidrule(r){2-3} \cmidrule(r){4-5} \cmidrule(r){6-7} \cmidrule(r){8-9} \cmidrule(r){10-11} \cmidrule(r){12-13}
& Acc~$\uparrow$ & Token~$\downarrow$ 
& Acc~$\uparrow$ & Token~$\downarrow$ 
& Acc~$\uparrow$ & Token~$\downarrow$ 
& Acc~$\uparrow$ & Token~$\downarrow$ 
& Acc~$\uparrow$ & Token~$\downarrow$ 
& \cellcolor{gray!20}Acc~$\uparrow$ & \cellcolor{gray!20}Token~$\downarrow$ \\
\midrule\midrule
\multicolumn{13}{l}{\textbf{Math-Shpherd}} \\
\cmidrule(r){1-13}
All-simple & 92.1 & 543.8 & 75.2 & 1092.6 & 50.0 & 1799.9 & 40.1 & 1124.9 & 10.0 & 1872.6 & 53.5 & 1286.8 \\
All-medium & \textbf{92.8} & 924.7 & 77.4 & 1915.0 & 47.5 & 2970.1 & 39.3 & 2086.4 & 13.3 & 3457.9 & 54.1 & 2270.8 \\
All-complex & 91.7 & 1707.5 & 74.2 & 3424.8 & 50.0 & 5318.1 & 40.1 & 3853.3 & 16.7 & 6535.5 & 54.5 & 4167.84 \\
\ding{72}~PATS (Ours) & 93.0 & 860.8 & 76.0 & 2216.5 & 57.5 & 3204.6 & 40.8 & 2020.1 & 10.0 & 5358.6 & 55.5 & 2732.1 \\
\midrule\midrule
\multicolumn{13}{l}{\textbf{Qwen2.5-Math-7B-PRM800K}} \\
\cmidrule(r){1-13}
All-simple & 93.8 & 585.5 & 77.2 & 1144.5 & 52.5 & 1729.5 & 40.1 & 1267.2 & 13.3 & 2055.7 & 55.4 & 1356.5 \\
All-medium & 94.1 & 1062.6 & 77.4 & 2244.2 & 60.0 & 3546.0 & 36.8 & 2527.2 & 13.3 & 4336.8 & 56.3 & 2743.4 \\
All-complex & 94.2 & 1916.0 & 79.8 & 4295.2 & 62.5 & 6556.1 & 41.9 & 4843.5 & 13.3 & 8768.7 & 58.3 & 5275.9 \\
\ding{72}~PATS (Ours) & 94.1 & 899.9 & 80.0 & 1954.1 & 55.0 & 3223.7 & 41.5 & 2071.8 & 16.7 & 3940.7 & 57.5 & 2418.0 \\
\bottomrule
\end{tabular}
\caption{All results of the Process Reward Model generalization experiments. Each policy model is evaluated on five math benchmarks. \textbf{All-simple}, \textbf{All-medium}, and \textbf{All-complex} perform fixed-mode reasoning with 2/4/8 candidates per step respectively. \textbf{PATS (Ours)} dynamically switches among modes at each step according to the PRM scores of the intermediate steps.}
\label{tab:prm-generalization}
\end{table*}

\end{document}